\documentclass{article} 
\usepackage[preprint]{colm2025_conference}

\usepackage{microtype}
\usepackage{hyperref}
\usepackage{url}
\usepackage{booktabs}
\usepackage{xcolor}
\usepackage{multirow}
\usepackage{colortbl}
\usepackage[export]{adjustbox}
\usepackage{array}
\usepackage{siunitx}
\usepackage{threeparttable}
\usepackage{cleveref}
\usepackage{wrapfig}
\usepackage{ulem}
\usepackage{lineno}
\usepackage{listings}  
\usepackage[utf8]{inputenc}
\usepackage[T1]{fontenc}

\definecolor{codegreen}{rgb}{0,0.6,0}
\definecolor{codegray}{rgb}{0.5,0.5,0.5}
\definecolor{codepurple}{rgb}{0.58,0,0.82}
\definecolor{backcolour}{rgb}{0.95,0.95,0.92}

\lstdefinestyle{mystyle}{
    backgroundcolor=\color{backcolour},   
    commentstyle=\color{codegreen},
    keywordstyle=\color{magenta},
    numberstyle=\tiny\color{codegray},
    stringstyle=\color{codepurple},
    basicstyle=\ttfamily\footnotesize,
    breakatwhitespace=false,         
    breaklines=true,                 
    captionpos=b,                    
    keepspaces=true,                 
    numbers=left,                    
    numbersep=5pt,                  
    showspaces=false,                
    showstringspaces=false,
    showtabs=false,                  
    tabsize=2
}

\definecolor{TableSeparator}{RGB}{220, 226, 250}

\definecolor{darkblue}{rgb}{0, 0, 0.5}
\hypersetup{colorlinks=true, citecolor=darkblue, linkcolor=darkblue, urlcolor=darkblue}

\title{%
  Biomed-Enriched: A Biomedical Dataset Enriched with LLMs\\
  for Pretraining and Extracting Rare and Hidden Content
  \\[1.5em]
  \centering
  \includegraphics[width=0.5\textwidth]{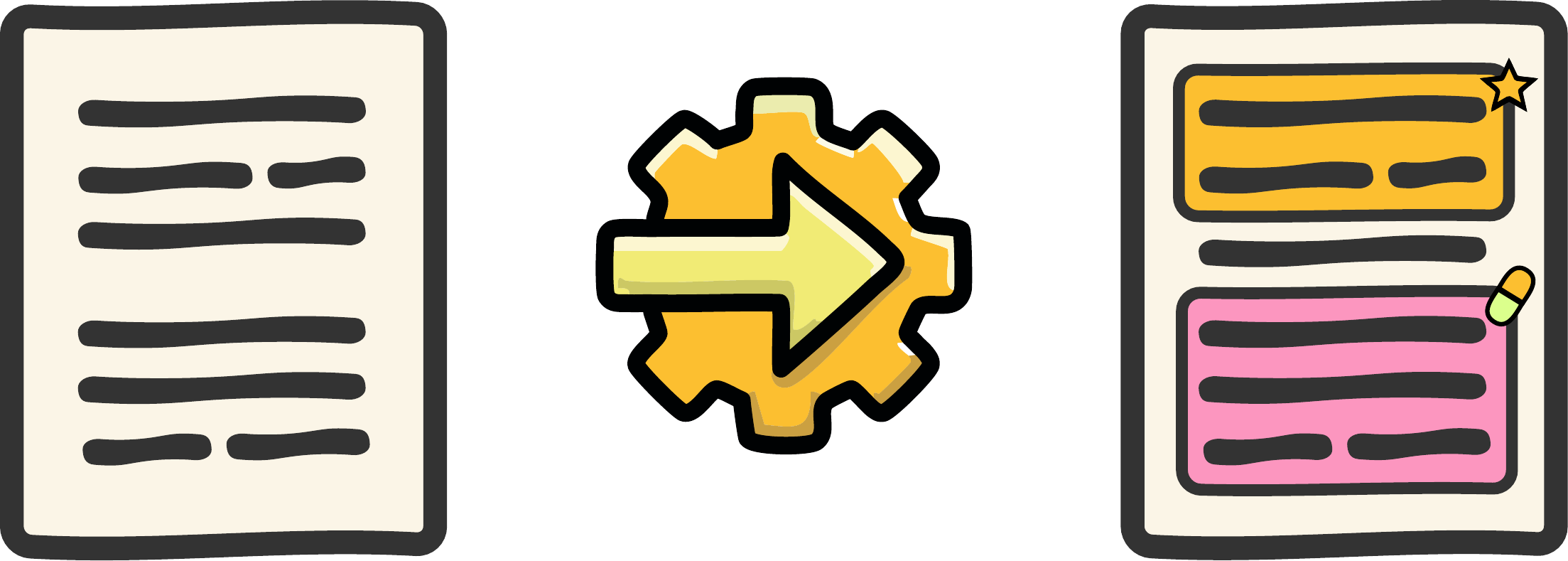}%
}

\author{%
  Rian Touchent, Nathan Godey \& Eric de la Clergerie\\
  Sorbonne Université\\
  INRIA Paris\\
  48 rue Barrault, Paris 75013, France\\
  \texttt{\{firstname.lastname\}@inria.fr}
}

\begin{document}

\ifcolmsubmission
\linenumbers
\fi

\maketitle

\begin{center}
  \href{https://huggingface.co/datasets/almanach/Biomed-Enriched}{%
    \includegraphics[height=1em,valign=c]{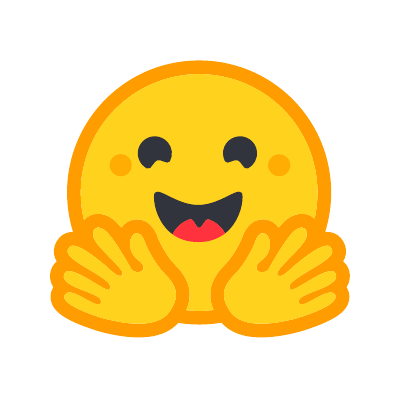}%
    \hspace{0.3em}%
    \texttt{almanach/Biomed-Enriched}%
  }
\end{center}
\vspace{1.5em}

\begin{abstract}
We introduce Biomed-Enriched, a biomedical text dataset constructed from PubMed via a two-stage annotation process. In the first stage, a large language model annotates 400K paragraphs from PubMed scientific articles, assigning scores for their type (review, study, clinical case, other), domain (clinical, biomedical, other), and educational quality. The educational quality score (rated 1 to 5) estimates how useful a paragraph is for college-level learning. These annotations are then used to fine-tune a small language model, which propagates the labels across the full PMC-OA corpus. The resulting metadata allows us to extract refined subsets, including 2M clinical case paragraphs with over 450K high-quality ones from articles with commercial-use licenses, and to construct several variants via quality filtering and domain upsampling. Clinical text is typically difficult to access due to privacy constraints, as hospital records cannot be publicly shared. Hence, our dataset provides an alternative large-scale, openly available collection of clinical cases from PubMed, making it a valuable resource for biomedical and clinical NLP. Preliminary continual-pretraining experiments with OLMo2 suggest these curated subsets enable targeted improvements, with clinical upsampling boosting performance by ~5\% on MMLU ProfMed and educational quality filtering improving MedQA and MedMCQA by ~1\%. Combinations of these techniques led to faster convergence, reaching same performance with a third of training tokens, indicating potential for more efficient and effective biomedical pretraining strategies.
\end{abstract}

\section{Introduction}

Large language models (LLMs) have demonstrated remarkable capabilities across a wide range of general tasks, from question answering to code generation. However, their performance often lags in specialized domains such as biomedical and clinical medicine, which demand domain expertise and precise terminology. This performance gap can be explained by the composition of standard pre-training corpora, which predominantly consist of web-scraped content from CommonCrawl. While diverse, these datasets lack sufficient representation of specialized knowledge required for complex biomedical reasoning. Although pre-training datasets are often supplemented with high-quality domain-specific corpora like PubMed, these curated additions represent only a small fraction compared to the vast amount of general web text \citep{li2024datacomp}. Available clinical text is particularly scarce in public datasets, with hospital records and clinical notes largely inaccessible due to strict privacy regulations. The situation is further complicated for non-English biomedical content, with resources like PMC containing over 98\% English articles. A central challenge in developing effective domain-specialized LLMs is therefore identifying strategies for curating, filtering, and upsampling domain-relevant documents to enhance performance on specialized tasks without compromising general capabilities.

\section{Related work}

To address the domain gap issue, researchers have employed continual pre-training on large biomedical corpora such as PubMed  and particularly its PMC Open Access Subset to enhance domain knowledge in LLMs. BioMistral \citep{labrak-etal-2024-biomistral}, for instance, underwent continual pre-training on 3 billion tokens from the PMC Open Access Subset, while Meditron \citep{chen2023meditron} fine-tuned Llama-2 on 46 billion tokens comprising PubMed abstracts, full papers, a general domain replay dataset, and clinical guidelines. Similarly, PMCLlama \citep{wu2024pmcllama} processed 75 billion tokens from PMC Open Access and medical textbooks, achieving significant improvements on biomedical benchmarks.

However, this process is compute-intensive for a moderate increase in performance. Meditron-70B \citep{chen2023meditron} required 128 A100 GPUs for 332 hours to achieve an average accuracy improvement of 1.8 percentage points on biomedical benchmarks, while BioMistral-7B \citep{labrak-etal-2024-biomistral} used 32 A100 GPUs for 20 hours, resulting in a 0.9-point performance decrease initially, but a 2.9-point improvement when using ensembling with different merging techniques with the original model.

PMC Open Access contains significant diversity and heterogeneity. Researchers typically employ filtering and upsampling strategies to better control training data composition. For instance, BioMistral \citep{labrak-etal-2024-biomistral} noted that 98.75\% of PMC Open Access articles are in English, leading them to upsample non-English articles. Meditron \citep{chen2023meditron} focused on high-quality, clinically relevant research by scoring articles (0–1) using MeSH tags, publication type, journal reputation, recency, and citation count. They filtered out low-scoring content while increasing the representation of higher-scoring articles. These methods, however, operate at the article level, overlooking valuable information in predominantly English articles that contain sections in other languages, or in articles that might be deemed low-quality overall but contain specific high-value passages.

More sophisticated filtering approaches have emerged to enhance pre-training data quality. While basic heuristic filtering using rules and perplexity scores from small language models trained on wikipedia showed improvements in language modeling, LLM-based semantic quality filtering has proven substantially more effective. FineWeb-Edu \citep{penedo2024fineweb} demonstrated the efficacy of model-based filtering by using Llama-3-70B-Instruct to annotate 500K documents from the FineWeb corpus based on educational value on a scale of 1 to 5. They then trained a smaller BERT-like model on these annotations and applied it to the entire FineWeb corpus, filtering out samples with scores below 3. Despite removing 92\% of the initial dataset, this refined subset outperformed both the complete FineWeb corpus and other open web datasets on knowledge-intensive benchmarks like MMLU \citep{hendrycks2021mmlu}, ARC \citep{clark2018arc}, and OpenBookQA \citep{mihaylov2018openbookqa}.

WebOrganizer \citep{wettig2025weborganizer} takes a complementary approach by organizing web content into structured taxonomies based on both topic and format. Rather than focusing solely on quality metrics, it unpacks monolithic web corpora into well-defined categories by distilling annotations from large language models into efficient classifiers. This systematic organization enables more refined data mixing strategies that improve model performance on downstream tasks. Importantly, their work demonstrates that domain-based organization provides valuable complementary benefits to quality-based filtering methods, as the two approaches can be combined to further enhance performance.

In our work, we develop a more refined approach to biomedical dataset curation through Biomed-Enriched, which applies LLM-driven annotation at the paragraph level rather than the document level. Building on techniques from FineWeb-Edu \citep{penedo2024fineweb} and WebOrganizer \citep{wettig2025weborganizer}, we focus on biomedical content from PMC Open Access, creating rich metadata about paragraph type, domain, educational quality, and language. This fine-grained approach enables more meaningful filtering and upsampling strategies that capture valuable information overlooked by article-level methods, including high-quality passages within lower-quality articles and non-English segments in predominantly English publications. We use a two-stage annotation process based on a smaller follow-up model to efficiently process the entire corpus. The detailed annotation allows us to extract valuable subsets particularly content related to clinical cases typically restricted due to privacy concerns. Our experiments show that this targeted data curation substantially improves efficiency in biomedical pre-training, resulting in faster convergence and targeted enhanced performance on domain-specific tasks.

Our contributions can be summarized in the following points:
\begin{itemize}
    \item We introduce Biomed-Enriched, a biomedical dataset created through a two-stage annotation process that enables fine-grained extraction of high-value content subsets from PubMed.
    \item We provide a large-scale collection of openly available clinical cases (2M paragraphs, including 450K high-quality ones), addressing a critical gap in accessible clinical text resources typically restricted by privacy regulations.
    \item We demonstrate targeted performance improvements through domain-specific upsampling, with clinical content upsampling yielding ~5\% gains on MMLU ProfMed and educational quality filtering of paragraphs improving medical QA tasks by ~1\%.
    \item We establish that strategically combining quality filtering and domain upsampling significantly improves data efficiency and targeted model performance, achieving equivalent scores using only one-third of the training tokens compared to standard approaches.
\end{itemize}

\section{Method}

We present Biomed-Enriched, a biomedical text dataset for enhanced biomedical training constructed through paragraph-level annotation and filtering. Our approach addresses the limitations of existing article-level filtering strategies by enabling more granular selection of high-value content. This is particularly relevant for clinical text which is traditionally difficult to access due to privacy constraints.

\subsection{Data Collection and Preprocessing}

We extracted text from the PubMed Central (PMC) Open Access Subset \citep{pmc_open_access}, containing approximately 4.5 million full-text scientific articles. This corpus, while valuable, presents challenges including heterogeneous quality, predominance of English content ($\approx98\%$), uneven representation of clinical cases, and variable educational value. Using a custom pipeline, we processed the raw XML files to extract article content, segment articles into 133M individual paragraphs, filter out non-textual elements, and retain only paragraphs containing a minimum of 64 tokens.

\subsection{Two-Stage Annotation Framework}

Our annotation process employed a two-stage approach. First, we used Llama-3.1-70B-Instruct \citep{touvron2024llama3} to annotate a diverse subset of 400,000 paragraphs from the PMC Open Access corpus across multiple dimensions: type classification, domain categorization, educational quality, and language identification. The annotation prompt instructed the model to assess each paragraph independently with reasoning for each rating (the complete prompt is provided in Appendix~\ref{appendix:prompt}).

To scale annotation to the full corpus, we distilled the LLM annotations into a smaller XLM-RoBERTa-base model \citep{conneau2020xlmr} trained to jointly predict all annotation dimensions using a standard multi-task learning framework. The distilled model achieved strong performance with 0.805 F1 score for domain classification, 0.854 F1 score for document type classification, and 0.245 MSE on educational quality score prediction, enabling efficient annotation of the entire corpus.

\begin{table}[t]
\begin{center}
\begin{tabular}{lp{0.75\columnwidth}}
\toprule
\textbf{Category} & \textbf{Description} \\
\midrule
\multicolumn{2}{l}{\textit{Document Type:}} \\
Clinical Case & Detailed report of symptoms, diagnosis, treatment, and follow-up of individual patients \\
Study & Research paragraph with methods, results, and discussion of experiments or observations \\
Review & Summary or synthesis of current knowledge on a specific topic \\
Other & Content not fitting above categories (editorials, commentaries, policy paragraphs) \\
\midrule
\multicolumn{2}{l}{\textit{Domain:}} \\
Clinical & Content relating to patient care, clinical trials, case reports, or practice guidelines \\
Biomedical & Scientific aspects of medicine and biology \\
Other & Content mentioning biomedical topics but focusing on administrative, policy, or general communications \\
\midrule
\multicolumn{2}{l}{\textit{Educational Quality:}} \\
Score 1 & Basic information relevant to biomedical topics, may contain irrelevant content \\
Score 2 & Addresses biomedical education elements but with limitations in coherence or depth \\
Score 3 & Appropriate for college-level curricula, introduces key concepts with reasonable coherence \\
Score 4 & Highly relevant educational content with clear writing style, minimal irrelevant information \\
Score 5 & Outstanding educational value, detailed reasoning with profound insights for college-level learning \\
\bottomrule
\end{tabular}
\end{center}
\caption{Classification used in our annotation framework. Document types categorize the structure and purpose of the content, domains identify the subject area focus, and educational quality scores assess pedagogical value for college-level biomedical learning on a scale from 1 (minimal value) to 5 (exceptional value).}
\label{tab:annotation-dimensions}
\end{table}

\subsection{Dataset Construction and Filtering}

Using the distilled model, we annotated the full corpus and constructed several dataset variants through strategic filtering and upsampling:

\textbf{BE-Base:} The complete unmodified PMC Open Access Subset serving as baseline.

\textbf{BE-Educational:} Preserves all articles but removes paragraphs with educational quality scores below 3.

\textbf{BE-Clinical:} Replicates articles with predominantly clinical domain content 10× in the training mix.

\textbf{BE-ClinicalCase:} Replicates articles containing at least one clinical case paragraph 10× to increase exposure to clinical narratives.

\textbf{BE-Prefix:} Prefixes each paragraph with its predicted annotations to allow modeling of metadata-content relationships.

\textbf{BE-French:} Upsamples articles containing French text 10× to address language imbalance.

\textbf{BE-All:} Combines quality filtering ($score \geq 3$), upsampling of clinical content, French text, and clinical cases, plus metadata prefixing.

For all variants, we preserved the original article structure. To maintain the contextual relationships between paragraphs within scientific articles, we employed an 8K context window during pre-training. This approach ensures that models can process complete scientific articles, allowing them to capture dependencies where information presented in earlier paragraphs is essential for properly understanding later content.

\begin{figure}[t]
\begin{center}
\includegraphics[width=1\columnwidth]{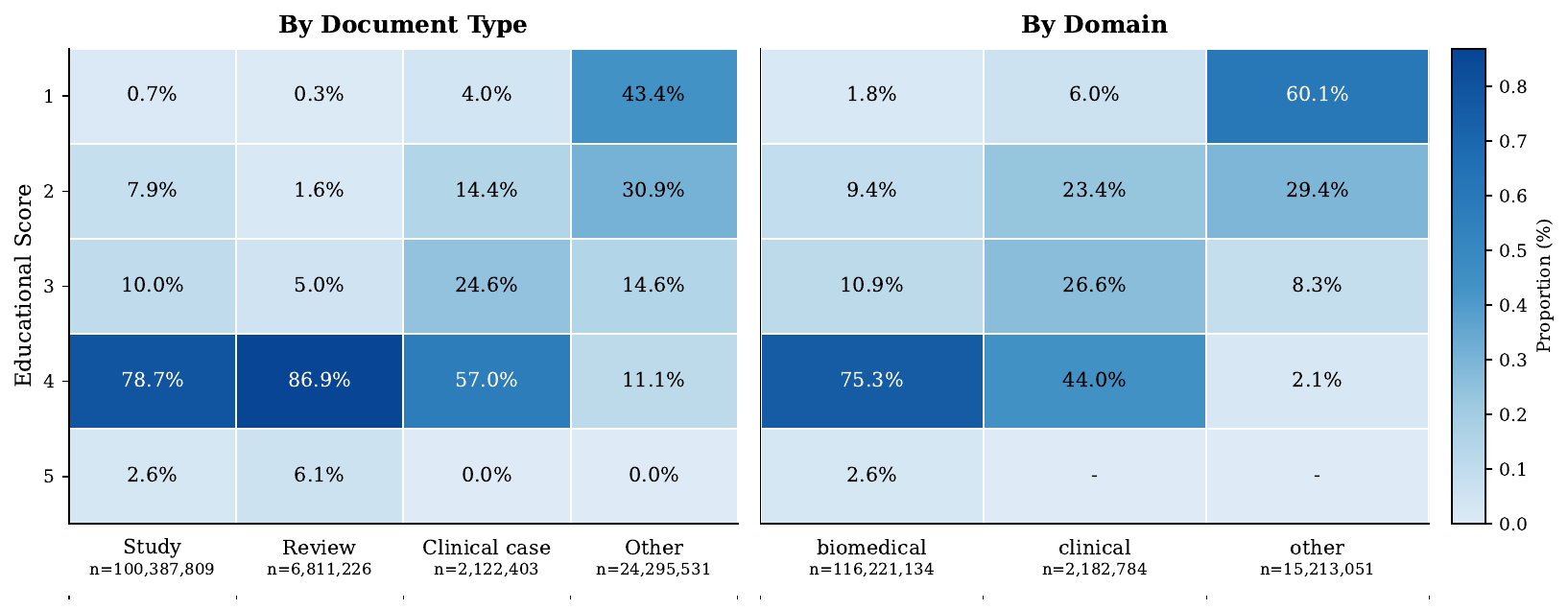}
\end{center}
\caption{Distribution of educational quality scores by document type and domain. Reviews and studies show the highest proportion of high scores, while clinical texts display more variance.}
\end{figure}

\begin{figure}[t]
\begin{center}
\includegraphics[width=1\columnwidth]{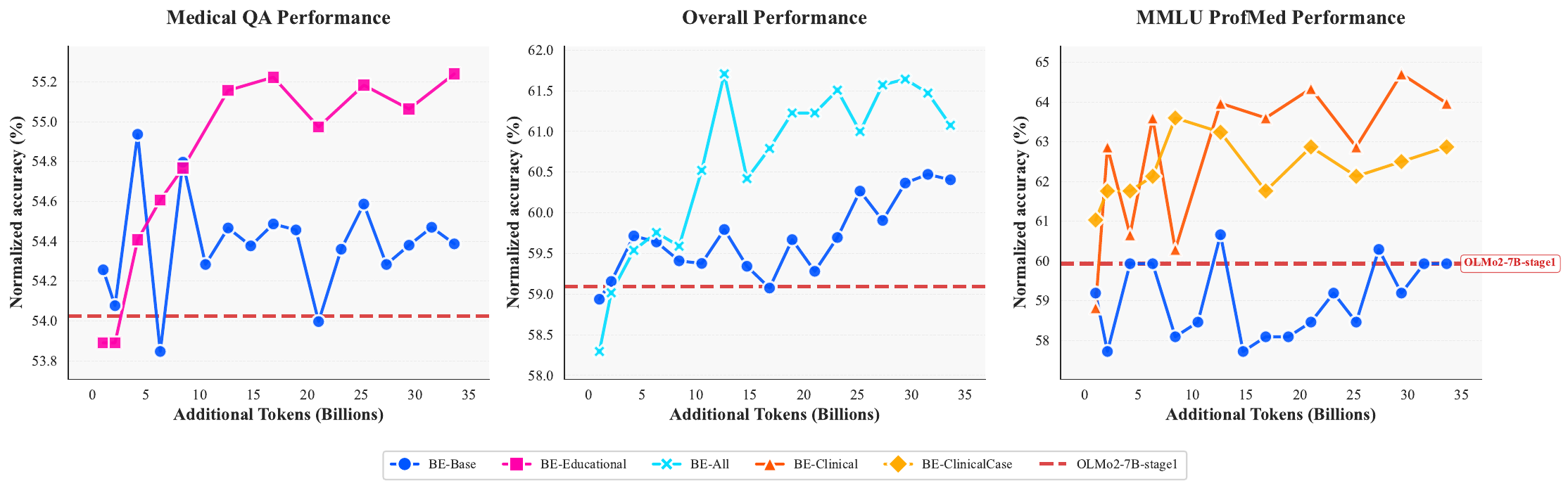}
\end{center}
\caption{Performance comparison across dataset variants showing training progression. BE-All achieves target performance with approximately one-third of the training tokens required by BE-Base.}
\end{figure}

\section{Data Analysis}


\paragraph{Score distribution.} The majority of PubMed paragraphs are rated with an educational score of 4, with a mean of 3.48 and a median of 4.00. This indicates a strong skew toward moderately high-quality educational content, which supports the feasibility of filtering based on score.

\paragraph{Score by document type.} Reviews and studies are the richest sources of educational content: 86.9\% of review paragraphs and 78.7\% of study paragraphs score a 4. Clinical cases, while less dominant, still show a significant share of high scores (57.0\% rated 4).

\paragraph{Score by domain.} Paragraphs tagged as biomedical are more likely to have high educational value (75.3\% score 4), while clinical texts show more variance (44.0\% score 4). In contrast, paragraphs labeled "other" rarely reach high scores (only 2.1\% rated 4), validating their exclusion in BE-Educational and BE-All.

\begin{wrapfigure}{r}{0.5\textwidth}
  \centering
  \includegraphics[width=0.9\linewidth]{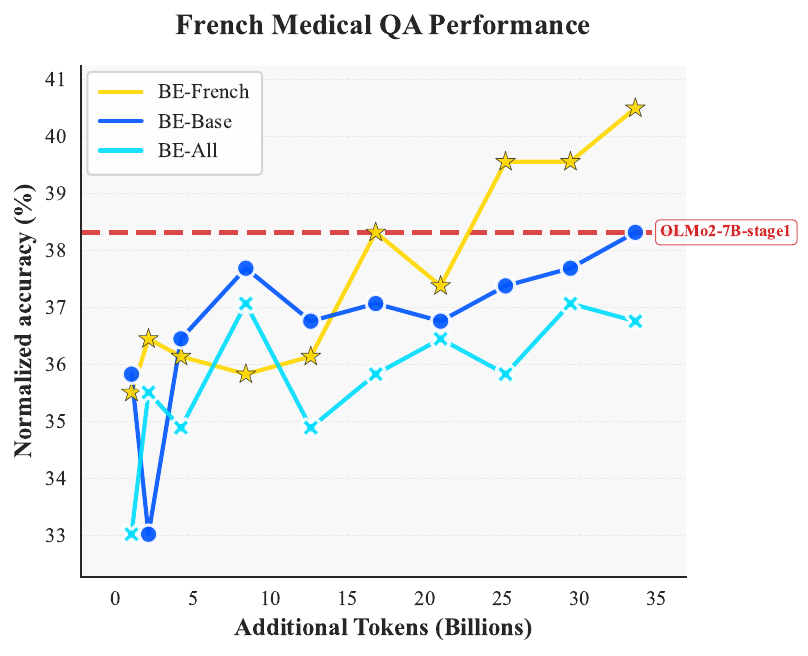}
  \caption{Performance on FrenchMedMCQA showing BE-French outperforming other variants, demonstrating effective language-specific improvement.}
  \label{fig:french-performance}
\end{wrapfigure}

\paragraph{Implications for filtering.} These distributions justify the score threshold (score $\geq$ 3) used in BE-Educational and BE-All, enabling targeted retention of higher-quality educational content while discarding noisy or low-value segments. The correlation between domain or type and educational score also explains why combining filters (as in BE-All) leads to consistent gains across tasks.

\subsection{Continual Pre-training}

Continual pre-training served as a method to evaluate the relevance and utility of our annotations. Our evaluation focuses on isolating the effects of data curation rather than pursuing state-of-the-art scores on benchmarks. A more powerful foundation model would likely yield higher absolute scores but would obscure the precise impact of our dataset. We selected OLMo2-7B-stage1 \citep{olmo2025furious} as our foundation model for continual pre-training, strategically choosing this intermediate checkpoint to better isolate the impact of our data curation techniques. While stage 1 has already developed strong language modeling capabilities, it precedes the knowledge-intensive tuning of stage 2, providing an ideal balance of baseline capabilities without the risk of catastrophic forgetting of instruction-following abilities during domain adaptation. Notably, the data mix used in phase 1 includes DCLM \citep{li2024datacomp}, which is a dataset obtained by filtering web-data using a classifier trained on instruct-data. Hence, OLMo2-7B already has relatively strong question-answering capabilities after stage 1.

Each Biomed-Enriched variant was trained for exactly 33.6 billion tokens using identical hyperparameters shown in Table \ref{tab:hyperparams}. We follow the annealing strategy of OLMo2 \citep{olmo2025furious} used in the mid-training phase. By maintaining strict parameter parity across experiments, we created a controlled environment focused solely on measuring the effectiveness of different data curation strategies.

\begin{table}[t]
\begin{center}
\begin{tabular}{ll}
\toprule
\textbf{Parameter} & \textbf{Value} \\
\midrule
Peak learning rate & 6.15e-5 \\
Minimal LR & 6.15e-6 \\
LR Decay & Linear \\
Batch size & 1024 \\
Weight decay & 0.1 \\
Context length & 8,192 tokens \\
Hardware & 128 MI250X GPUs \\
Training time (hours / GPU-hours) & 68 / 8700 \\
\bottomrule
\end{tabular}
\end{center}
\caption{Hyperparameters used for continual pre-training}\label{tab:hyperparams}
\end{table}

\subsection{Evaluation Framework}

We evaluated how our annotation-guided corpus refinement affects model performance during continual pre-training. We measured performance at regular intervals throughout training on several biomedical benchmarks to understand how effectively models acquire domain knowledge from differently curated datasets.

Our evaluation consists of zero-shot testing on MMLU medical subcategories, MedQA, MedMCQA, and PubMedQA. For the French adaptation assessment, we used 5-shot evaluation on FrenchMedMCQA \citep{labrak-etal-2022}. We compared our models against three baselines: OLMo2-7B-stage1, Llama-3.1-8B, and Meditron-70B.

Beyond final performance, we specifically analyzed the token-efficiency of each approach
. This metric helps identify which annotation dimensions and filtering strategies provide the most valuable training signal for biomedical knowledge acquisition, offering practical insights for efficient domain adaptation.

\begin{table*}[t]
\centering
\normalsize
\resizebox{\columnwidth}{!}{
\begin{threeparttable}[b]
  \centering
  \sisetup{detect-weight=true}
    \begin{tabular}{lS[table-format=2.2]S[table-format=2.2]S[table-format=2.2]S[table-format=2.2]S[table-format=2.2]S[table-format=2.2]S[table-format=2.2]S[table-format=2.2]S[table-format=2.2]S[table-format=2.2]}
    \toprule
    \multicolumn{1}{c}{\multirow{3}[6]{*}{\textbf{}}} & \multicolumn{3}{c}{\textbf{Medical QA}} & \multicolumn{6}{c}{\textbf{MMLU Medical}} & \multicolumn{1}{c}{} \\
    \cmidrule(lr){2-4}  \cmidrule(lr){5-10}
     & \textbf{MedQA} & \textbf{MedMCQA} & \textbf{PubMedQA} & \textbf{Anat} & \textbf{Clin} & \textbf{Bio} & \textbf{Med} & \textbf{Gen} & \textbf{Prof} & \textbf{Avg} \\
    \midrule
    \rowcolor{TableSeparator} \multicolumn{11}{c}{\rule{0pt}{10pt}\textbf{SOTA models (for reference)}} \\[1pt]
    \multicolumn{1}{p{7.585em}}{Llama-3-8B} & 59.70 & 57.47 & 74.80 & 68.89 & 74.72 & 78.47 & 61.85 & 83.00 & 70.22 & 69.90 \\
    \multicolumn{1}{p{7.585em}}{Meditron-70B} & 57.10 & 46.80 & 76.60 & 53.30 & 66.70 & 76.30 & 63.00 & 69.00 & 71.60 & 64.49 \\
    \rowcolor{TableSeparator} \multicolumn{11}{c}{\rule{0pt}{10pt}\textbf{Benchmark Results by Dataset Variant}} \\[1pt]
    \noalign{\vspace{2pt}}
    \multicolumn{1}{p{7.585em}}{\small{OLMo2-7B-stage1}} & 45.33 & 41.14 & 75.60 & 54.81 & 63.40 & 69.44 & 53.18 & 69.00 & 59.93 & 59.09 \\
    BE-Base & 44.85 & 41.91 & 76.40 & \underline{57.04} & 64.15 & \bfseries 70.83 & \bfseries 59.54 & \underline{69.00} & 59.93 & \underline{60.41} \\
    \arrayrulecolor{gray!40}\midrule[1pt]\arrayrulecolor{black}
    BE-Clinical & 41.95 & 39.35 & 76.60 & 53.33 & 63.40 & 65.28 & 58.38 & 66.00 & \bfseries 63.97 & 58.70 \\
    BE-ClinicalCase & 42.11 & 39.52 & 76.60 & \underline{57.04} & 64.91 & 66.67 & \bfseries 59.54 & \underline{69.00} & \underline{62.87} & 59.81 \\
    BE-Prefix & \underline{45.72} & 41.76 & \bfseries 77.80 & \underline{57.04} & 64.53 & \underline{68.75} & 57.23 & 66.00 & 61.76 & 60.07 \\
    BE-Educational & 45.64 & \bfseries 43.08 & \underline{77.00} & \underline{57.04} & \underline{65.28} & 68.06 & 56.65 & \bfseries 71.00 & 58.82 & 60.29 \\
    BE-All & \bfseries 47.21 & \underline{42.79} & 76.60 & \bfseries 60.00 & \bfseries 65.66 & 68.06 & 58.96 & \underline{69.00} & 61.40 & \bfseries{61.08} \\
    \bottomrule
    \end{tabular}%
    \begin{tablenotes}
\centering
      \small
      \item Note: MMLU abbreviations: Anat=Anatomy, Clin=Clinical Knowledge, Bio=College Biology, Med=College Medicine, Gen=Medical Genetics, Prof=Professional Medicine.
\end{tablenotes}
\end{threeparttable}
}
\caption{Comprehensive performance results across medical benchmarks for different dataset enrichment strategies.}
\label{tab:comprehensive}%
\end{table*}

\section{Results}

\paragraph{Overall performance.} BE-All achieved the highest average performance across benchmarks at 61.08\%, surpassing BE-Base (60.41\%) by a small but consistent margin (+0.67 pts, Table \ref{tab:comprehensive}). Its strongest improvements appeared in MedQA (47.21\%), MMLU Anatomy (60.00\%), and Clinical Knowledge (65.66\%), suggesting the effectiveness of combining multiple targeted enrichment strategies.

\paragraph{Clinical enrichment.} BE-Clinical significantly boosted performance on MMLU Professional Medicine benchmark (63.97\%, +4.04 pts vs. BE-Base, Figure 2). This improvement was stable from early training, highlighting how clinical narratives enhance the model’s clinical reasoning abilities efficiently.

\paragraph{Educational filtering.} BE-Educational consistently improved performance on medical question-answering tasks, notably Medical Genetics (71.00\%, +2 pts), MedMCQA (43.08\%, +1.17 pts), and PubMedQA (77.00\%, +0.6 pts). These tasks likely benefit from the knowledge present in educationally high-quality paragraphs (Figure 2).

\paragraph{Metadata prefixing.} BE-Prefix
specifically improved performance on PubMedQA (77.80\%, +1.4 pts vs. BE-Base). Providing explicit paragraph-level metadata helped primarily with structured document comprehension, but it had limited benefits for other tasks.

\paragraph{General biomedical knowledge trade-off.} BE-Base performed better on College Biology (70.83\%) than enriched variants. Building a biology variant (BE-Bio) could be an interesting future direction, as the current dataset does not specifically target this domain.

\paragraph{Non-English enrichment.} BE-French showed clear improvements in French medical QA (FrenchMedMCQA), achieving 40.5\% accuracy, significantly surpassing BE-Base and the OLMo2-7B-stage1 baseline (38.32\%, Figure 1). These results illustrate effective adaptation to non-English contexts using only upsampling of annotated paragraphs which could be applied to other languages.

\paragraph{Data efficiency and training stability.} As shown in Figure 2, BE-All reached robust benchmark performance using roughly one-third of the tokens required by BE-Base. Individual enrichments (Educational, Clinical) also displayed early and stable improvements, underscoring potential reductions in training time and computational cost.

\section{Discussion}

Our findings demonstrate that strategic data enrichment substantially improves training efficiency and targeted model effectiveness in biomedical pretraining.

\paragraph{Benefits of paragraph-level annotation.}  Paragraph-level annotation allows us to identify specific types of content, such as clinical case descriptions, that may only appear in isolated sections of scientific articles. This makes it possible to collect and focus on content that is otherwise difficult to target at scale, like the 2 million clinical case paragraphs we were able to extract through this approach.

\paragraph{Task-specific enrichment strategies.} Our results highlight clear task-specific benefits: educational filtering supports knowledge-intensive QA, clinical upsampling improves clinical reasoning, and metadata enrichment enhances structured comprehension. This demonstrates the potential of targeted pretraining strategies, adaptable to other domains or annotations, to better serve specific use cases, beyond aiming for general-purpose models trained with massive compute.

\paragraph{Non-English generalization.} The successful adaptation via targeted French upsampling highlights our approach's flexibility and effectiveness for non-English scenarios without additional model complexity or extensive computational overhead, which can be applied to other languages.

\paragraph{Combined vs. targeted strategies.} Combining multiple enrichments (BE-All) generally produced the highest scores, but individual enrichments provide critical insights into their respective contributions and trade-offs. This emphasizes the value of selecting enrichments strategically based on specific tasks rather than relying exclusively on combined approaches.

\paragraph{Limitations and trade-offs.} Our study has some limitations worth noting. Our experiments used relatively small language models (7B parameters), and larger models might produce different or clearer patterns. Specialized enrichment strategies may slightly reduce broader biomedical knowledge, indicating the need to balance targeted knowledge enhancement against overall knowledge retention. Future studies should explicitly address these comparisons and explore optimal blends of detailed and general-domain knowledge enrichment to further improve biomedical NLP models.

\section{Conclusion}

We introduced Biomed-Enriched, a dataset created by annotating the PMC Open Access corpus at the paragraph level. This design enabled the discovery of valuable content, such as clinical case narratives and high-quality educational passages, that is often difficult to isolate through broader filtering methods. Our continual pretraining experiments show improved stability and data efficiency, reaching comparable performance using only a third of the training tokens.

While the combined strategy (BE-All) performs best overall, each targeted enrichment shows distinct strengths, demonstrating how aligning data selection with task requirements can yield meaningful improvements. This highlights the potential of fine-grained, annotation-driven curation to support more focused and efficient pretraining.

Rather than aiming for generic models trained on as much data as possible, our findings support a shift toward modular, adaptable strategies. The annotation pipeline used here can be extended to new domains, tasks, or languages—offering a flexible foundation for building specialized models that meet the evolving needs of the biomedical NLP community.

\section*{Acknowledgments}
This work was granted access to the HPC resources of IDRIS under the allocation 2024-AD011014393R1 made by GENCI.

\section*{Ethics Statement}

This work focuses on the creation and use of biomedical datasets derived from publicly available scientific literature. All data used in this study come from the PubMed Central Open Access Subset, which is explicitly licensed for text and data mining. No private or patient-identifiable clinical data were used or accessed.

\bibliography{colm2025_conference}
\bibliographystyle{colm2025_conference}

\appendix
\section{Appendix}

\subsection{LLM Annotation Prompt}
\label{appendix:prompt}

Below is the prompt used to instruct Llama-3.1-70B-Instruct for the first stage of our annotation process:

\begin{lstlisting}
Below is an extract from a scientific article. Evaluate whether the extract has a high educational value and could be useful in an educational setting for teaching at the college level in biomedical sciences using the additive 5-point scoring system described below. 

Points are accumulated based on the satisfaction of each criterion: 
- Add 1 point if the extract provides some basic information relevant to biomedical topics, even if it includes some irrelevant or non-academic content like advertisements and promotional material.
- Add another point if the extract addresses certain elements pertinent to biomedical education but does not align closely with academic standards. It might mix educational content with non-educational material, offering a superficial overview of potentially useful topics, or presenting information in a disorganized manner and incoherent writing style.
- Award a third point if the extract is appropriate for educational use and introduces key concepts relevant to college-level biomedical curricula. It is coherent though it may not be comprehensive or could include some extraneous information. It may resemble an introductory section of a textbook or a basic tutorial that is suitable for learning but has notable limitations like treating concepts that are too complex for introductory students.
- Grant a fourth point if the extract is highly relevant and beneficial for educational purposes at the college level, exhibiting a clear and consistent writing style. It could be similar to a chapter from a textbook or a tutorial, offering substantial educational content, including exercises and solutions, with minimal irrelevant information, and the concepts are appropriate for college students. The content is coherent, focused, and valuable for structured learning.
- Bestow a fifth point if the extract is outstanding in its educational value, perfectly suited for teaching at the college level in biomedical sciences. It follows detailed reasoning, the writing style is easy to follow and offers profound and thorough insights into the subject matter, devoid of any non-educational or overly complex content.

Based on these factors, give a score from 1 to 5.

Also, classify the relevant domain as either "biomedical", "clinical", or "other" following the guidelines provided below:
- Clinical: Extract appears to be written in a clinical context by a healthcare professional. It should contain information directly related to patient care, such as details from clinical trials, case reports, or clinical guidelines.
- Biomedical: Extract contains substantive information on biomedical sciences. It could be from a research paper or textbook, focusing on the scientific aspects of medicine and biology.
- Other: Extract mentions biomedical or clinical topics but doesn't provide substantive content in these areas. This category includes:
  1. Administrative or funding information about biomedical research
  2. General news or public communications about medical topics
  3. Policy discussions related to healthcare
  4. Any content that talks about biomedical or clinical subjects without providing actual scientific or medical information

Additionally, identify the type of document, this category includes:
1. Clinical case: A detailed report of the symptoms, signs, diagnosis, treatment, follow-up, etc. of an individual patient.
2. Study: Research-based document that includes methods, results, and discussions about experiments or observations, often involving multiple subjects or data points.
3. Review: A document that summarizes or evaluates the current state of knowledge on a specific topic.
4. Other: Any other type of document not fitting the above categories.

After examining the extract:
- Briefly justify your quality classification, up to 100 words on one line using the format: "Explanation: <justification>"
- Conclude with the quality classification using the format: "Educational score: <classification>"
- Conclude with the domain classification using the format: "Domain: <classification>"
- Conclude with the document type classification using the format: "Document type: <classification>"
\end{lstlisting}

\end{document}